\documentclass[11pt]{article}
\usepackage{acl2017}
\usepackage{times}
\usepackage{url}
\usepackage{latexsym}
\usepackage{xspace}
\usepackage{booktabs}
\usepackage{color}
\usepackage{graphicx}
\usepackage{tabulary}
\usepackage{enumitem}
\usepackage{url}
\usepackage{amsfonts}

\newenvironment{tightitemize}%
  {\begin{itemize}[topsep=0pt, partopsep=0pt] %
    \setlength{\itemsep}{0pt}%
    \setlength{\parskip}{0pt}%
    }%
  {\end{itemize}}
  
  \usepackage{color}
  {\begin{enumerate}≈ \footnotesize%
    \setlength{\itemsep}{0pt}%
    \setlength{\parskip}{0pt}%
    }%
  {\end{enumerate}}

\makeatletter
\newcommand{\@emptybiblabel}[1]{}
\makeatother

\usepackage{microtype}
\usepackage{multirow}
\usepackage{verbatim}
\usepackage{amsmath,amsthm,amssymb}
\usepackage{array}
\usepackage[scaled=0.86]{helvet}
\usepackage{ifthen}
\usepackage{courier}
\usepackage[linesnumbered,vlined,ruled]{algorithm2e}
\aclfinalcopy

\belowdisplayskip \abovedisplayskip

\newcommand{\sts}{{{\textsc{Seq2Seq}}}\xspace}

\newcommand{\todo}[1]{\textcolor{red}{#1}}

\title{Adversarial Learning for Neural Dialogue Generation}
\author{
Jiwei Li\hspace{1pt}$^{{{\bf 1}}}$, Will Monroe\hspace{1pt}$^{{{\bf 1}}}$, Tianlin Shi\hspace{1pt}$^{{{\bf 1}}}$,
S\'ebastien Jean$^2$,  Alan Ritter\hspace{1pt}$^{{{\bf 3}}}$ and Dan Jurafsky\hspace{1pt}$^{{{\bf 1}}}$ 
\\[0.4cm]
$^1$Stanford University, Stanford, CA, USA\\
$^2$New York University, NY, USA \\
$^3$Ohio State University, OH, USA \\
{\tt jiweil,wmonroe4,tianlins,jurafsky@stanford.edu} \\
{\tt sebastien@cs.nyu.edu}\\
{\tt  ritter.1492@osu.edu} 
}

\begin{document}
\maketitle

\begin{abstract}
In this paper, drawing intuition from the Turing test, we propose using adversarial training for open-domain dialogue generation: the system is trained to produce sequences
that are
 indistinguishable from human-generated dialogue utterances. We cast the task as a reinforcement learning (RL) problem where we jointly train two systems, a generative model to produce response sequences, and a discriminator---analagous to the human evaluator in the Turing test--- to distinguish between 
 the 
 human-generated dialogues and the machine-generated ones. The outputs from the discriminator are 
 then
 used as rewards for the generative model, pushing the system to generate dialogues that mostly resemble human dialogues.

In addition to adversarial training we describe a model
for adversarial {\em evaluation} that 
uses success in fooling an adversary as a dialogue evaluation metric,
while avoiding a number of potential pitfalls.
Experimental results on several
metrics, including adversarial evaluation, demonstrate
that the adversarially-trained system generates higher-quality responses
than previous baselines. 
\end{abstract}

\section{Introduction}
\label{sec:intro}
Open domain dialogue generation  \cite{ritter2011data,sordoni2015neural,xu2016incorporating,wen2016network,li2016persona,serban2016multiresolution,serban2016hierarchical} 
aims at generating meaningful and coherent dialogue responses given the dialogue history.
Prior systems, e.g., phrase-based machine translation systems \cite{ritter2011data,sordoni2015neural} or end-to-end neural systems \cite{shang2015neural,vinyals2015neural,li2015diversity,yao2015attention,luan2016lstm}
 approximate such a goal by 
predicting the next dialogue utterance  given the dialogue history using the maximum likelihood estimation (MLE) objective. 
Despite its success, this over-simplified training objective leads to problems:
responses are dull, generic 
 \cite{sordoni2015neural,serban2015hierarchical,li2015diversity}, 
 repetitive, and short-sighted \cite{li2016deep}. 

Solutions to these problems require answering a few fundamental questions: 
what are the crucial aspects that characterize an ideal conversation, how can we quantitatively measure them, and how can we incorporate them into 
a machine learning system? 
For example, 
\newcite{li2016deep} manually define three ideal dialogue properties (ease of answering, informativeness and coherence) and use a reinforcement-learning framework to train the model to generate highly rewarded responses. 
\newcite{yu2016strategy} use keyword retrieval confidence as a reward. 
However, 
it is widely
acknowledged that  
manually defined reward functions can't possibly
cover all crucial aspects and can lead to suboptimal generated utterances. 

A good dialogue model should generate utterances indistinguishable from human dialogues.
Such a goal suggests a training objective 
resembling the idea of the Turing test \cite{turing1950computing}.
We borrow the idea of adversarial training \cite{goodfellow2014generative,denton2015deep} in 
computer
vision, in which we jointly train two models, 
a generator (a neural  \sts model) that defines the probability of generating a dialogue sequence, and 
a discriminator
that labels dialogues as human-generated or machine-generated. 
This discriminator  is analogous to  the evaluator in the Turing test.
We cast the task as a reinforcement learning problem, in which the quality of machine-generated utterances is measured by its ability to fool the discriminator into believing that it is a human-generated one. The output from the discriminator is used as a reward to the generator, pushing it to generate 
utterances indistinguishable from human-generated dialogues. 

The idea of a Turing test---employing an evaluator 
to distinguish  machine-generated texts from human-generated
ones---can be applied not only to training but also testing,
where it goes by the name of adversarial evaluation. Adversarial
evaluation was first employed in \newcite{bowman2015generating} to evaluate sentence generation quality, and preliminarily studied for dialogue generation by \newcite{kannan}. 
In this paper, we discuss potential pitfalls of adversarial evaluations and 
  necessary steps to avoid them and make evaluation reliable.

Experimental results demonstrate that our approach
produces more interactive, interesting, and non-repetitive responses than standard
\sts models trained using the MLE objective function.

\section{Related Work}
\paragraph{Dialogue generation}
Response generation for dialogue can be viewed as 
 a source-to-target transduction problem.
\newcite{ritter2011data}
frame the generation problem as a machine translation problem.  
\newcite{sordoni2015neural}
improved Ritter et al.'s system 
by rescoring
the outputs of a phrasal MT-based conversation
system with a neural model incorporating
prior context.
Recent progress in \sts models have inspired several efforts \cite{vinyals2015neural,serban2015hierarchical,serban2016,luan2016lstm} to  build end-to-end conversational systems that first apply an encoder to map a message to a distributed vector representing its meaning and then generate a response from the vector.

Our work adapts the encoder-decoder model to RL training, and can thus be viewed as an extension of
 \newcite{li2016deep}, but with more general RL rewards.  \newcite{li2016deep} 
simulate dialogues
between two virtual agents, using policy gradient
methods to reward sequences that display
three useful conversational properties: informativity,
coherence, and ease of answering. 
 Our work is also
related to recent efforts to
 integrate the \sts and reinforcement learning  paradigms, drawing on the advantages of both \cite{wen2016network}.
For example, \newcite{Su}  combine reinforcement learning with neural generation on tasks with real users.
\newcite{RE} train an end-to-end RL dialogue model using  human users.  

Dialogue quality is traditionally evaluated \cite[e.g.]{sordoni2015neural} using word-overlap metrics such as BLEU and METEOR scores used for machine translation.  
 Some recent work \cite{liu2016not} has started to look at more flexible and reliable evaluation metrics 
 such as human-rating prediction \cite{hey} and next utterance classification \cite{lowe2016evaluation}.

\paragraph{Adversarial networks} 
The idea of generative adversarial networks has enjoyed great success in computer vision \cite{radford2015unsupervised,chen2016infogan,salimans2016improved}. 
Training is formalized as a game in which the generative model is trained to generate outputs to fool the discriminator; the technique has been successfully applied to image generation.

However, to the best of our knowledge, this idea has not achieved comparable success in NLP. 
This is due to the fact that unlike in vision, text generation is discrete, which makes the error
outputted
 from the discriminator hard to backpropagate to the generator. 
Some recent work has begun to address this issue:
\newcite{lamb2016professor} propose providing the discriminator with the intermediate hidden vectors of the generator rather than its sequence outputs. Such a strategy  makes the system differentiable and achieves promising results in tasks like character-level language modeling and handwriting generation. 
\newcite{yu2016seqgan}  use policy gradient reinforcement learning to backpropagate the error from the discriminator, showing improvement in multiple generation tasks such as poem generation,  speech language generation
and music generation. 
Outside of sequence generation,
\newcite{chen2016adversarial} apply the idea of adversarial training to sentiment analysis and \newcite{zhang2017aspect} apply the idea to domain adaptation tasks.

 Our work is distantly related to recent work that formalizes sequence generation as an action-taking problem in reinforcement learning. \newcite{ranzato2015sequence}
 train RNN decoders in a \sts model using policy gradient
to obtain competitive machine translation results.
\newcite{bahdanau2016actor} take this a step further by training an actor-critic RL model for machine translation.
Also related is recent work \cite{shen2015minimum,wiseman2016sequence} to address the  issues of exposure bias and
loss-evaluation mismatch in neural translation.

\section{Adversarial Training for Dialogue Generation}
In this section, we describe in detail the components of the proposed adversarial reinforcement learning model. 
The problem can be framed as follows: given a dialogue history $x$ consisting of a sequence of dialogue utterances,\footnote{We approximate the dialogue history using the concatenation of two preceding utterances. We found that using more than 2 context utterances yields very tiny performance improvements for \sts models.} the model needs to generate a response $y=\{y_1,y_2,...,y_T\}$.
 We view the 
 process of sentence generation 
   as a sequence of actions that are taken according to a policy defined by an
encoder-decoder recurrent neural network.

\subsection{Adversarial REINFORCE}
The adversarial REINFORCE algorithm consists of two components:
a generative model $G$ and a discriminative model $D$.

\paragraph{Generative model} The generative model $G$ defines the
 policy that generates a response $y$ given dialogue history $x$. 
It takes a  form similar to \sts models, which first map the source input to a vector representation using a recurrent net
and then compute the probability of generating each token in the target using a softmax function.

\paragraph{Discriminative model} The discriminative model $D$ is a binary classifier that takes as input a sequence of dialogue utterances $\{x,y\}$
 and outputs a label indicating whether the input is generated by humans or machines.  
The input dialogue is encoded into a vector representation 
using a hierarchical encoder  \cite{li2015hierarchical,serban2016building},\footnote{To be specific, each utterance $p$ or $q$ is mapped to a vector representation $h_p$ or $h_q$ using LSTM \cite{hochreiter1997long}.
Another LSTM is put on sentence level, mapping the context dialogue sequence to a single representation.} which is then fed to a 2-class softmax function, returning the probability of the input dialogue episode being a machine-generated dialogue
(denoted $Q_-(\{x,y\})$)
 or a human-generated dialogue  (denoted $Q_+(\{x,y\})$). 
\paragraph{Policy Gradient Training}
 The key idea of the system is to encourage the generator to generate utterances that are indistinguishable from human generated dialogues. We use policy gradient methods to achieve such a goal, in which the 
score of current utterances being human-generated ones assigned by the discriminator
(i.e.,  $Q_+(\{x,y\})$)
 is used as a reward for the generator, which is trained to maximize the expected reward of generated utterance(s) using the REINFORCE algorithm \cite{williams1992simple}:
 \begin{equation}
 J(\theta)=\mathbb{E}_{y\sim p(y|x)}(Q_+(\{x,y\})|\theta)
 \label{lb1}
 \end{equation}
Given the input dialogue history $x$, the  bot generates a dialogue utterance $y$ by sampling from the policy. 
The concatenation of the generated utterance $y$ and the input $x$ is fed to the discriminator. 
 The gradient of \eqref{lb1} is approximated using the likelihood ratio trick \cite{williams1992simple,glynn1990likelihood,aleksand1968stochastic}:
 \begin{multline}
\nabla J(\theta)\approx [Q_+(\{x,y\})-b(\{x,y\})] \\ \nabla\log \pi(y|x)
\\ = [Q_+(\{x,y\})-b(\{x,y\})] \\ \nabla\sum_t \log p(y_t|x,y_{1:t-1})
\label{partial}
\end{multline}
where $\pi$ denotes the probability of the generated responses. 
 $b(\{x,y\})$ denotes the baseline value to reduce the variance of the estimate while keeping it unbiased.\footnote{
 Like \newcite{ranzato2015sequence}, 
 we train another neural network model (the critic) to estimate the value (or future reward) of current state (i.e., the dialogue history) under the current policy $\pi$. The critic network takes as input the dialogue history, transforms it to a vector representation using a hierarchical network and maps the representation to a scalar. The network is optimized based on the mean squared loss between the estimated reward and the real reward.} The discriminator is simultaneously updated with the human generated dialogue that contains dialogue history $x$ as a positive example and the machine-generated dialogue as a negative example. 
\subsection{Reward for Every Generation Step (REGS)}
The  REINFORCE algorithm
described has the disadvantage that the expectation of the reward is approximated by only one sample, and the reward associated with this sample 
(i.e.,  $[Q_+(\{x,y\})-b(\{x,y\})]$ in Eq\eqref{partial})
is used for all actions (the generation of each token) in the generated sequence. 
Suppose, for example, the input history is {\it what's your name}, the human-generated response is {\it I am John}, and the machine-generated response is {\it I don't know}. 
The vanilla REINFORCE model assigns the same negative reward to all tokens within the human-generated response (i.e., {\it I}, {\it don't}, {\it know}), whereas 
proper credit assignment in training would  give separate rewards, most likely a neutral reward for the token {\it I}, and negative rewards to {\it don't} and {\it know}. 
We call this {\it reward for every generation step}, abbreviated {\it REGS}.

Rewards for intermediate steps or partially decoded sequences are thus necessary. Unfortunately, the discriminator is trained to assign scores to fully generated sequences, but not partially decoded ones. 
We propose two strategies for computing intermediate step rewards by (1) using Monte Carlo (MC) search and (2) training a discriminator that is able to assign rewards to partially decoded sequences.

In (1) Monte Carlo search, given a partially decoded $s_P$, the model 
keeps sampling tokens from the distribution
 until the decoding finishes. Such a process is repeated $N$ (set to 5) times and the $N$ generated   sequences will share  a common prefix $s_P$.
These $N$ sequences 
are fed to the discriminator, the average score of which is used as a reward for the $s_P$. A similar strategy is adopted in \newcite{yu2016seqgan}. 
The downside of MC is that 
it requires repeating the sampling process for each prefix of each sequence  and is thus significantly time-consuming.\footnote{Consider one target sequence with length 20, we need to sample 5*20=100 full sequences to get rewards for all intermediate steps. Training one batch with 128 examples roughly takes roughly 1 min on a single GPU, which is computationally intractable considering the size of the dialogue data we have. We thus parallelize the sampling processes, distributing jobs across 8 GPUs. }

In (2), we directly train a discriminator that is able to assign rewards to both fully and partially decoded sequences. 
We break 
the generated sequences into partial sequences, namely 
$\{y^+_{1:t}\}_{t=1}^{N_{Y^+}}$ and $\{y^-_{1:t}\}_{t=1}^{N_{Y^-}}$
 and use 
 all instances in  $\{y^+_{1:t}\}_{t=1}^{N_{Y^+}}$ as positive examples and  instances  $\{y^-_{1:t}\}_{t=1}^{N_{Y^-}}$ as negative examples. 
The problem with such a strategy is that earlier actions in a sequence are 
shared among multiple training examples for
 the discriminator 
(for example, token $y^+_1$ is contained in all partially generated sequences, which results in overfitting. 
To mitigate this problem, 
we adopt a  strategy similar to when training value networks in  {\it AlphaGo} \cite{silver2016mastering}, in which 
for each collection of subsequences of $Y$, we randomly sample only one example from $\{y^+_{1:t}\}_{t=1}^{N_{Y^+}}$ and one example from $\{y^-_{1:t}\}_{t=1}^{N_{Y^-}}$, which are treated as positive and negative examples
to update the discriminator. 
Compared with the Monte Carlo search model, this strategy is significantly more time-effective, but comes with the weakness that the discriminator becomes less accurate  after partially decoded sequences are added in as training examples. 
We find that the MC model performs better when training time is less of an issue.

For each partially-generated sequence $Y_t=y_{1:t}$, the discriminator gives a classification score $Q_+(x,Y_t)$. 
We compute the baseline $b(x,Y_t)$
using a similar model  to the vanilla REINFORCE model. 
This yields the following gradient to update the generator:
 \begin{multline}
\nabla J(\theta)\approx \sum_t  (Q_+(x,Y_t)-b(x,Y_t))  \\
\nabla\log p(y_t|x,Y_{1:t-1})
\label{action}
\end{multline}
Comparing \eqref{action} with \eqref{partial}, we can see that
the values for  
rewards and baselines are different among generated tokens in the same response. 

\paragraph{Teacher Forcing}
Practically, we find that  updating the generative model only using Eq.~\ref{lb1} leads to  unstable training for both vanilla Reinforce and REGS, with the perplexity value skyrocketing after training the model for a few hours (even when the generator is initialized using a pre-trained \sts model). The reason this happens is that the generative model can only be indirectly exposed to the gold-standard target sequences through the reward 
passed back from the
 discriminator, and this reward is used to promote or discourage its (the generator's) own generated sequences. Such a training strategy is fragile: once the generator (accidentally) deteriorates in some training batches and the discriminator consequently does an extremely good job in recognizing sequences from the generator,  
the generator immediately gets lost. It knows that its generated sequences are bad based on the rewards outputted from the discriminator, but it does not know what sequences are good and how to push itself to generate these good sequences (the odds of generating a good response from random sampling are minute, due to the vast size of the space of possible sequences). Loss of the reward signal leads to a breakdown in the training process.

To alleviate this issue and give the generator more direct access to the gold-standard targets, 
we propose also feeding human generated responses to the generator for model updates.  
The most straightforward strategy is for the discriminator to automatically assign a reward of 1 (or other positive values) to the human generated responses and for
the generator to use this reward to update itself on human generated examples. 
This can be seen as having a teacher intervene with the generator some fraction of the time and force it to generate the true responses, 
an approach that is similar to the professor-forcing algorithm of \newcite{lamb2016professor}.

A closer look reveals that 
this modification is the same as the standard training of \sts models, making the final training
alternately update the \sts model using the adversarial objective and the MLE objective. One can think of the 
professor-forcing 
model as a regularizer to 
regulate the generator once it starts deviating from the training dataset.

We also propose another workaround, in which the discriminator first assigns a reward to a human generated example using its own model,  and the generator then updates itself using this reward on the human generated example only if the reward is larger than the baseline value. Such a strategy has the advantage that different weights for model updates are assigned to different human generated examples (in the form of different reward values produced by the generator) and that human generated examples are always associated with non-negative weights. 

A sketch of the proposed model is shown in Figure \ref{fig:adver-reinforce}. 
\begin{figure}
\small
\line(1,0){220} \\
{\bf For} number of training iterations {\bf do} \\
.~\hspace{0.3cm}{\bf For} i=1,D-steps {\bf do} \\
.~\hspace{0.8cm}Sample (X,Y) from real data \\
.~\hspace{0.8cm}Sample $\hat{Y}\sim G(\cdot|X)$\\
.~\hspace{0.8cm} Update $D$ using $(X,Y)$ as positive examples and $(X,\hat{Y})$ as negative examples. \\
.~\hspace{0.3cm}{\bf End} \\
.\\
.~\hspace{0.2cm} {\bf For} i=1,G-steps {\bf do} \\
.~\hspace{0.8cm}Sample (X,Y) from real data \\
.~\hspace{0.8cm}Sample $\hat{Y}\sim G(\cdot|X)$ \\
.~\hspace{0.8cm}Compute Reward $r$ for $(X,\hat{Y})$ using $D$.\\
.~\hspace{0.8cm}Update $G$  on $(X,\hat{Y})$ using reward $r$\\
.~\hspace{0.8cm}Teacher-Forcing: Update $G$  on $(X,Y)$\\
.~\hspace{0.3cm}{\bf End} \\
{\bf End} \\
\line(1,0){220}
\caption{A brief review of the proposed adversarial reinforcement algorithm
for training the generator $G$ and discriminator $D$.
The reward $r$ from the discriminator $D$ can be computed using different strategies according to whether using REINFORCE or REGS.  
The update of the generator $G$ on $(X,\hat{Y})$ can be done by either using Eq.\ref{partial} or Eq.\ref{action}.
D-steps is set to 5 and G-steps is set to 1.}
\label{fig:adver-reinforce}
\end{figure}

\subsection{Training Details}
We first  pre-train the generative model by predicting target sequences given the dialogue history. We trained a  \sts model
  \cite{sutskever2014sequence}   
 with an attention mechanism \cite{bahdanau2014neural,luong2015effective}  on the OpenSubtitles dataset. 
 We followed protocols recommended by \newcite{sutskever2014sequence}, such as gradient clipping, mini-batch and learning rate decay.
We also pre-train the discriminator. To generate negative examples, 
we  decode   part of the training data. 
Half of the negative examples are generated using beam-search with mutual information reranking 
as described in \newcite{li2015diversity},
and the other half is generated from sampling.

For 
data processing, 
model training and decoding  (both the proposed adversarial training model and the standard \sts models), we employ a few  strategies that  improve response quality, including: 
(2) Remove training examples with length of responses shorter than a threshold (set to 5).
We find that this significantly improves the general response quality.\footnote{To compensate for the loss of short responses, one can train a separate model using short sequences.}
(2) Instead of using the same learning rate for all examples, using a weighted learning rate that considers the average tf-idf score for tokens within the response.
Such a strategy
 decreases the influence from dull and generic utterances.\footnote{We treat each sentence as a document. Stop words are removed.
 Learning rates are normalized within one batch. 
 For example, suppose $t_1$, $t_2$, ..., $t_i$, ... ,$t_N$ denote the tf-idf scores for  sentences within current batch and $lr$ denotes the original learning rate. The learning rate for sentence with index $i$ is $N\cdot lr\cdot \frac{t_i}{\sum_{i'}t_{i'}}$. 
 To avoid exploding learning rates for sequences with extremely rare words, the tf-idf score of a sentence 
 is capped at $L$ times the minimum tf-idf score in the current batch. $L$ is empirically chosen and is set to 3.
  } (3) Penalizing intra-sibling ranking when doing beam search decoding to promote N-best list diversity as described in \newcite{li2016simple}.  (4) Penalizing word types (stop words excluded) that have already been generated. Such a strategy dramatically decreases the rate of repetitive responses such as {\it no. no. no. no. no.} or contradictory responses such as {\it I don't like oranges but i like oranges}. 

\section{Adversarial Evaluation}
In this section, we discuss strategies for successful adversarial evaluation. 
Note that the proposed adversarial training and adversarial evaluation 
are separate procedures. 
They are independent of each other and share no common parameters.

 The idea of adversarial evaluation, first proposed by 
\newcite{bowman2015generating}, is to 
train a discriminant
function
 to separate  generated and
true sentences, in an attempt to evaluate the model's sentence generation capability.
The idea has been preliminarily studied by \newcite{kannan} in the context of dialogue generation. 
Adversarial evaluation  also resembles the idea of the Turing test,
which requires a human evaluator 
to distinguish  machine-generated texts from human-generated ones. 
Since it is time-consuming and costly to 
 ask a human to talk to a model 
and give judgements,
we  train a machine evaluator
in place of the  human evaluator to distinguish the human dialogues and machine dialogues, and we use it to measure 
 the general quality of the generated responses. 

Adversarial evaluation involves both  training and testing. 
At training time, the evaluator is trained to
label dialogues as machine-generated (negative) or  human-generated (positive). 
At test time, the trained evaluator is evaluated on a held-out dataset.
If the human-generated dialogues and machine-generated ones are indistinguishable, the model  will achieve 50 percent accuracy at test time.  
\subsection{Adversarial Success}
We define Adversarial Success ({\it AdverSuc} for short) to be the fraction of instances in which a model is capable of fooling the evaluator. {\it AdverSuc} is the difference between 1 and the accuracy 
achieved by the  evaluator. Higher values of {\it AdverSuc} for a dialogue generation model are better. 
\subsection{Testing the Evaluator's Ability}
One caveat  with the adversarial evaluation methods is that 
they are model-dependent. 
We approximate the human evaluator in the Turing test with an automatic evaluator and assume that the evaluator is perfect: low accuracy of the discriminator should indicate high quality of the responses, since we interpret this to mean the generated responses are indistinguishable from the human ones. Unfortunately, there is another factor that can lead to low discriminative accuracy: a poor discriminative model. 
 Consider a  discriminator that always gives random labels or always gives the same label.
 Such an evaluator 
   always yields a high {\it AdverSuc} value of 0.5. 
   \newcite{bowman2015generating} propose  two different discriminator models 
separately using {\it unigram} features and {\it neural} features. It is hard to tell which feature set is more reliable. 
The standard strategy of testing the model on a held-out development set is not suited to this case, since a model that overfits the development set is necessarily superior. 

To deal with this issue, we propose setting up a few manually-invented situations to test the ability of the automatic evaluator.
 This is akin to setting up  examinations to test the ability of the human evaluator in the Turing test. 
We  report not only the {\it AdverSuc} values,  but also the scores that the evaluator
achieves
 in these manually-designed test cases, indicating how much we can trust the reported {\it AdverSuc}. 
 We develop scenarios in which we know  in advance how a perfect evaluator should behave, and then compare   {\it AdverSuc}  from a discriminative model with the gold-standard  {\it AdverSuc}. Scenarios we design include:
 \begin{tightitemize}
 \item We use human-generated dialogues as both positive examples and negative examples. A perfect evaluator should give an {\it AdverSuc} of $0.5$ (accuracy $50\%$), which is the gold-standard result.
  \item We use machine-generated dialogues as both positive examples and negative examples. A perfect evaluator should give an {\it AdverSuc} of $0.5$ (accuracy $50\%$).
 \item We use original human-generated dialogues as positive examples and dialogues 
 consisting of
  random utterances 
  as negative examples. A perfect evaluator should give an {\it AdverSuc} of $0$ (accuracy $100\%$).
 \item We use original human-generated dialogues as positive examples and dialogues with 
 responses replaced by the utterances immediately following the true responses. 
 A perfect evaluation should give an {\it AdverSuc} of $0$ (accuracy $100\%$).
\end{tightitemize}
The {\it evaluator reliability error} (ERE) is the average deviation of an evaluator's adversarial error
 from the gold-standard error in the above tasks, with equal weight for each task. 
The smaller the error, the more reliable the evaluator is. 

\begin{table*}[t]
\centering
\footnotesize
\begin{tabular}{p{4cm}p{11cm}}\hline
{\bf Input}&tell me ... how long have you had this falling sickness ?\\
{\bf ~~Vanilla-MLE}&i 'm not a doctor .\\
{\bf ~~Vanilla-Sample}&well everything you did was totally untrue .\\
{\bf ~~REINFORCE}&i don 't know how long it 's been .\\
{\bf ~~REGS Monte Carlo}& A few months, I guess .\\\hline
{\bf Input}&so i had the doctors test sammy 's response to conditioning .\\
{\bf ~~Vanilla-MLE}&sammy wrote the test sammy wrote the test .\\
{\bf ~~Vanilla-Sample}&objects pick objects up objects objects objects objects objects \\
{\bf ~~REINFORCE}&i 'm not sure that 's relevant .\\
{\bf ~~REGS Monte Carlo}&so he took the pills .\\\hline
{\bf Input}&what are you going to do ?\\
{\bf ~~Vanilla-MLE}&i 'm going to the movies .\\
{\bf ~~Vanilla-Sample}&get him outta here first !\\
{\bf ~~REINFORCE}&i 'm going to get you .\\
{\bf ~~REGS Monte Carlo}&i 'm going to get you a new car .\\\hline
{\bf Input}&they fear your power your intellect .\\
{\bf ~~Vanilla-MLE}&you 're the only one who knows what 's going on .\\
{\bf ~~Vanilla-Sample}&when they are conquered and you surrender they will control all of us .\\
{\bf ~~REINFORCE}&i 'm afraid i 'm not ready yet .\\
{\bf ~~REGS Monte Carlo}&i 'm not afraid of your power .\\\hline
\end{tabular}
\caption{Sampled responses from different models. More in Appendix Tables \ref{example-appendix1} and \ref{example-appendix2}.}
\end{table*}

\subsection{Machine-vs-Random Accuracy}
Evaluator reliability error uses scenarios constructed from human-generated
dialogues to assess feature or hyper-parameter choice for the evaluator. Unfortunately, no machine-generated responses are involved in the ERE metric. 
The following example illustrates the serious weakness resulting from this strategy:
as will be shown in the experiment section, 
when inputs are decoded using greedy or beam search models,  
most generation systems to date yield an adversarial success less than 10 percent (evaluator accuracy 90 percent). 
But when using sampling for decoding, the adversarial success skyrockets to around 40 percent,\footnote{Similar results are also reported in \newcite{kannan}.} only 10 percent less than what's needed to pass the Turing test. 
A close look at the decoded sequences using sampling tells a different story: the responses from sampling are sometimes incoherent, irrelevant or even ungrammatical. 

We thus propose an additional sanity check, in which we report the accuracy of distinguishing between machine-generated responses and randomly sampled responses ({\it machine-vs-random} for short). 
This resembles the N-choose-1 metric described in \newcite{shao15}. 
Higher accuracy indicates that the generated responses are distinguishable from randomly sampled human responses, indicating that the generative model is not fooling the generator simply by introducing randomness. 
As we will show in Sec.~\ref{sec:experiments}, using sampling results in high {\it AdverSuc} values but low {\it machine-vs-random} accuracy. 

\section{Experimental Results} \label{sec:experiments}
In this section, 
we detail experimental results on adversarial success and human evaluation.  

\begin{table}[htb]
\centering
\small
\begin{tabular}{ccc}
Setting&ERE\\\hline
SVM+Unigram&0.232\\
Concat Neural &0.209 \\
Hierarchical Neural &0.193 \\
SVM+Neural+multil-features&0.152 \\
\hline
\end{tabular}
\caption{ERE scores obtained by different models.}
\label{ERE}
\end{table}

\subsection{Adversarial Evaluation}
\paragraph{ERE} We first test 
 adversarial evaluation models with different 
 feature sets and 
 model architectures for reliability, as measured by evaluator reliability error (ERE).
 We explore the following models:
   (1) {\it SVM+Unigram}:  SVM using unigram features.\footnote{Trained using the SVM-Light package \cite{joachims2002learning}.}
 A multi-utterance dialogue (i.e., input messages and responses) is transformed to a unigram representation; (2) 
{\it Concat Neural}: 
a neural classification model with 
a softmax function that takes as input the concatenation of representations of constituent dialogues sentences;
 (3) {\it Hierarchical Neural}: 
  a hierarchical encoder    
  with a  structure similar to the discriminator used in the reinforcement; and
  (4) 
  {\it SVM+Neural+multi-lex-features}: 
  a SVM model that uses the following features: unigrams,  neural representations of dialogues obtained by the neural model trained using strategy (3),\footnote{The representation before the softmax layer.} the forward likelihood $\log p(t|s)$ and backward likelihood $p(s|t)$.

ERE scores obtained by different models are reported in Table \ref{ERE}. 
As can be seen, the {\it hierarchical neural} evaluator (model 3) is more reliable than simply concatenating the sentence-level representations (model 2).
Using the combination of neural features and lexicalized features yields the most reliable evaluator. 
For the rest of this section, we report results obtained 
by the 
{\it Hierarchical Neural} setting due to its end-to-end nature, despite its inferiority to {\it SVM+Neural+multil-features}. 

Table \ref{adv} presents
 {\it AdverSuc} values for different models, along with {\it machine-vs-random} accuracy described in Section 4.3. 
Higher values of  {\it AdverSuc}  and  {\it machine-vs-random} are better. 

Baselines we consider include standard \sts models using greedy decoding ({\it MLE-greedy}), beam-search ({\it MLE+BS}) and sampling, as well as the 
mutual information reranking model of \newcite{li2015diversity} with two algorithmic variations: (1) MMI+$p(t|s)$, in which a large N-best list is first generated using a pre-trained \sts model and then reranked by the backward probability $p(s|t)$ and (2) MMI$-p(t)$, in which language model probability is penalized during decoding. 

Results are shown in Table \ref{adv}. What first stands out  is decoding using sampling (as discussed in Section 4.3), achieving a significantly higher {\it AdverSuc} number than all the rest models. 
However, this does not indicate the superiority of the sampling decoding model, since the {\it machine-vs-random} accuracy is at the same time significantly lower. This means that sampled responses based on \sts models are not only hard for an evaluator to distinguish from real human responses, but also from randomly sampled responses.
A similar, though much less extreme, effect is observed for MMI$-p(t)$, which has an {\it AdverSuc} value slightly higher than {\it Adver-Reinforce}, but a significantly lower {\it machine-vs-random} score. 

By comparing different baselines, we find that MMI+$p(t|s)$ is better than {\it MLE-greedy}, which is in turn better than {\it MLE+BS}. This result is in line with human-evaluation results from \newcite{li2015diversity}. 
The two proposed adversarial algorithms achieve better performance than the baselines. We expect this to be the case, since the adversarial algorithms are trained on an objective function 
more similar to the
evaluation metric (i.e., adversarial success). 
{\it REGS} performs slightly better than the vanilla REINFORCE algorithm. 

\begin{table}
\small
\centering
\begin{tabular}{ccc}
\hline
Model&{\it AdverSuc}&{\it machine-vs-random} \\\hline
MLE-BS&0.037&0.942 \\
MLE-Greedy&0.049&0.945 \\ 
MMI+$p(t|s)$&0.073&0.953\\
MMI-$p(t)$&0.090& 0.880\\
Sampling&0.372&0.679\\\hline
Adver-Reinforce&0.080&0.945 \\
Adver-REGS&0.098&0.952\\
\hline
\end{tabular}
\caption{{\it AdverSuc} and {\it machine-vs-random} scores achieved by different training/decoding strategies.}
\label{adv}
\end{table}

\subsection{Human Evaluation}
For human evaluation, we follow protocols defined in \newcite{li2016deep}, employing
crowdsourced judges to evaluate a random sample of
200 items. We present both an input message and the
generated outputs to 3 judges and ask them to decide
which of the two outputs is better ({\it single-turn}
general quality). Ties are permitted. Identical
strings are assigned the same score. 
We also
 present the judges with {\it multi-turn}
conversations simulated between the two agents. Each conversation
consists of 3 turns.
Results are presented in Table \ref{human}.
We observe a significant quality improvement on both 
single-turn quality and multi-turn quality from the proposed adversarial model.
It is worth noting that the reinforcement learning system described in \newcite{li2016deep}, which 
simulates conversations between two bots and 
is trained based on manually designed reward functions, only improves multi-turn dialogue quality, while the model described in this paper improves both single-turn and multi-turn dialogue generation quality. 
This confirms that the reward adopted in adversarial training is more general, natural and effective in training dialogue systems.

\begin{table}
\small
\centering
\begin{tabular}{cccc}\\\hline
Setting &adver-win &adver-lose &tie\\\hline
single-turn& 0.62& 0.18 &0.20 \\
multi-turn& 0.72 &0.10& 0.18\\\hline
\end{tabular}
\caption{The gain
from the proposed adversarial model
 over the mutual information system
based on pairwise human judgments.}
\label{human}
\end{table}

\section{Conclusion and Future Work}
In this paper, drawing  intuitions from the Turing test, we propose using an adversarial training approach for response generation. We cast the model in the framework of reinforcement learning and train a generator based on the signal from a discriminator to generate response sequences indistinguishable from human-generated dialogues.
We observe clear performance improvements on multiple metrics from the adversarial training strategy.

The adversarial training model should theoretically benefit a variety of generation tasks in NLP. Unfortunately, in preliminary experiments applying the same training paradigm to machine translation, we did not observe a clear performance boost. We conjecture that this is because the adversarial training strategy 
is more beneficial to
 tasks in which there is 
 a big discrepancy between the distributions of the generated sequences and the  reference target sequences. In other words, the adversarial approach is more beneficial on tasks in which entropy of the targets is  high. Exploring this relationship further is a focus of our future work. 
 ~~\\~~\\
 \paragraph{Acknowledgements}
 The authors thank Michel Galley, Bill Dolan, Chris Brockett, Jianfeng Gao 
and other members of the NLP group at Microsoft Research, as well as
 Sumit Chopra and Marc'Aurelio Ranzato from Facebook AI Research for helpful discussions and comments. 
Jiwei Li is supported by a Facebook Fellowship,  which we gratefully acknowledge. 
This work is also partially supported by the NSF under award
IIS-1514268, and the DARPA Communicating
with Computers (CwC) program under ARO prime
contract no. W911NF- 15-1-0462, IIS-1464128. Any opinions,
findings, and conclusions or recommendations expressed
in this material are those of the authors and
do not necessarily reflect the views of DARPA, the NSF,
or Facebook.

\bibliographystyle{acl_natbib}
\bibliography{emnlp2016}

\begin{table*}[!htb]
\centering
\footnotesize
\begin{tabular}{p{4cm}p{8cm}}\hline
{\bf Input}&maybe his eyes were too big for his stomach .\\
{\bf Vanilla-MLE}&he 's got a big mouth .\\
{\bf Vanilla-Sample}&i 'm sure he 's unhappy with the fact .\\
{\bf Reinforce}&he 's a little bit of a pain .\\
{\bf MC}&i don 't think so .\\\hline
{\bf Input}&they didn 't have much success with this problem commander .\\
{\bf Vanilla-MLE}&they 're not the only ones who have been in the system .\\
{\bf Vanilla-Sample}&it 's the new priority only .\\
{\bf Reinforce}&they 're not the only ones who can help us .\\
{\bf MC}&can we find someone else ?\\\hline
{\bf Input}&and he carried them all the time wherever he went .\\
{\bf Vanilla-MLE}&i don 't know who he is .\\
{\bf Vanilla-Sample}&the church ... suffered the winds of 900 meters with horse aid .\\
{\bf Reinforce}&he was a good man .\\
{\bf MC}&he was a very good man .\\\hline
{\bf Input}&somebody probably flushed their pet snake down the toilet .\\
{\bf Vanilla-MLE}&i 'm sure the dog will be here soon .\\
{\bf Vanilla-Sample}&i could at least think .\\
{\bf Reinforce}&i don 't think so .\\
{\bf MC}&i don 't think so .\\\hline
{\bf Input}&excuse me but how did we pay for these suits ?\\
{\bf Vanilla-MLE}&i 'm not a beggar .\\
{\bf Vanilla-Sample}&is this the money eugene ?\\
{\bf Reinforce}&i don 't know what to do .\\
{\bf MC}& how much money you owe me ?\\\hline
{\bf Input}&... while the beautiful and happy world was madly destroyed ;\\
{\bf Vanilla-MLE}&i was born in the world of the world\\
{\bf Vanilla-Sample}&as in childhood i took you to one tenth of a postcard\\
{\bf Reinforce}&and the world was full of joy .\\
{\bf MC}&and the world was full of joy .\\\hline
{\bf Input}&i was kinda hoping that you were my dad .\\
{\bf Vanilla-MLE}&i 'm just a little bit of a little bit of a little bit of a little bit of a\\
{\bf Vanilla-Sample}&shiny gold gold and silver in the mountains .\\
{\bf Reinforce}&i 'm not gonna be a baby .\\
{\bf MC}&i 'm not going to be a father .\\\hline
{\bf Input}&the man who does that is not afraid to die .\\
{\bf Vanilla-MLE}&he 's not afraid of the dark .\\
{\bf Vanilla-Sample}&even the program is the only truth\\
{\bf Reinforce}&he 's also not afraid of me .\\
{\bf MC}&he 's such a brave man !\\\hline
{\bf Input}&do you have any idea what caused the explosion ?\\
{\bf Vanilla-MLE}&i think it 's the same thing .\\
{\bf Vanilla-Sample}&i 'm still very nervous .\\
{\bf Reinforce}&maybe it 's just a coincidence .\\
{\bf MC}&i don 't know if it 's a coincidence .\\\hline
{\bf Input}&don 't bother the citizens in hong kong island anymore\\
{\bf Vanilla-MLE}&i 'll be back in a few days\\
{\bf Vanilla-Sample}&you wiii be transferred to the target court\\
{\bf Reinforce}&i 'm not sure if you can handle it without bothering them .\\
{\bf MC}&i 'm not going to be a nuisance\\\hline
{\bf Input}&you know you ought to find yourself a girl .\\
{\bf Vanilla-MLE}&you know what i mean ?\\
{\bf Vanilla-Sample}&it 's full of crime ...\\
{\bf Reinforce}&i don 't know what you 're talking about .\\
{\bf MC}&i don 't want to be a girl .\\\hline
{\bf Input}&i got you the big screen tv deluxe UNknown machine\\
{\bf Vanilla-MLE}&i 'm gonna be a little late for the movie .\\
{\bf Vanilla-Sample}&you mean you 're living ?\\
{\bf Reinforce}&i 'm gonna get you a new one .\\
{\bf MC}&i 'm gonna get you a new tv set .\\\hline
{\bf Input}&we won 't give her the chance to surprise us .\\
{\bf Vanilla-MLE}&she 's not going to be a party member .\\
{\bf Vanilla-Sample}&the timing 's good and bad .\\
{\bf Reinforce}&we 'll have to get her back .\\
{\bf MC}&i don 't think we have to worry about her .\\\hline
{\bf Input}&does it live in this matter we call dna ?\\
{\bf Vanilla-MLE}&i don 't think so .\\
{\bf Vanilla-Sample}&and it requires time to be wrong and no one believes it .\\
{\bf Reinforce}&it 's a genetic disorder .\\
{\bf MC}&i don 't think so .\\\hline
\end{tabular}
\caption{Appendix: Responses sampled from different models.}
\label{example-appendix1}
\end{table*}

\begin{table*}[!htb]
\centering
\footnotesize
\begin{tabular}{p{4cm}p{8cm}}\hline

{\bf Input}&i 'll have to drive home .\\
{\bf Vanilla-MLE}&i 'll be back in a minute .\\
{\bf Vanilla-Sample}&let 's go for some fun .\\
{\bf Reinforce}&i 'll drive you home .\\
{\bf MC}&i don 't have a car .\\\hline
{\bf Input}&i had to work late and then i went for a walk .\\
{\bf Vanilla-MLE}&i was in the middle of the road .\\
{\bf Vanilla-Sample}&my body was just pounding .\\
{\bf Reinforce}&i was just trying to help you .\\
{\bf MC}&i was just trying to get a little sleep .\\\hline
{\bf Input}&lots of cream lots of sugar .\\
{\bf Vanilla-MLE}&i 'm gonna be in the kitchen .\\
{\bf Vanilla-Sample}&was this gay or red ?\\
{\bf Reinforce}&i don 't want to be a vegetable .\\
{\bf MC}&i 'm not a very good cook .\\\hline
{\bf Input}&so they just let him go ?\\
{\bf Vanilla-MLE}&he 's not gonna let them go through the motions .\\
{\bf Vanilla-Sample}&they asked him whether we could go see ernesto ?\\
{\bf Reinforce}&i don 't know what to do .\\
{\bf MC}&he 's not going to be here any longer .\\\hline
{\bf Input}&you must not bomb this town .\\
{\bf Vanilla-MLE}&you have to go back to the city .\\
{\bf Vanilla-Sample}&the one is a multiple choice .\\
{\bf Reinforce}&I think I have to .\\
{\bf MC}&I think I have to .\\\hline
{\bf Input}&communism recently failed in eastern europe .\\
{\bf Vanilla-MLE}&the war is not over .\\
{\bf Vanilla-Sample}&there are people who hope that in tokyo they won 't risk losing their departing .\\
{\bf Reinforce}&and the people who are not the same .\\
{\bf MC}&the war is still going on .\\\hline
{\bf Input}&you should stop in more often .\\
{\bf Vanilla-MLE}&I don 't think so .\\
{\bf Vanilla-Sample}&is it possible as old as you ?\\
{\bf Reinforce}&i 'm not going to be a burden on you .\\
{\bf MC}&i don 't want to be a burden .\\\hline{\bf Input}&he couldn 't have heard us .\\
{\bf Vanilla-MLE}&he was in the middle of the night  .\\
{\bf Vanilla-Sample}&he should have been somewhere out here .\\
{\bf Reinforce}&he 's not a good guy .\\
{\bf MC}&he 's got a lot of nerve.\\\hline
{\bf Input}&that 's our wake up call .\\
{\bf Vanilla-MLE}&we 're gonna be late for the meeting .\\
{\bf Vanilla-Sample}&c 'mon c 'mon let 's get outta here .\\
{\bf Reinforce}&i 'll be right back .\\
{\bf MC}&we 're gonna have to get to the station .\\\hline
{\bf Input}&i had 'em when i started .\\
{\bf Vanilla-MLE}&i 'm gonna get 'em .\\
{\bf Vanilla-Sample}&those weak rotten guys had me in a rush on a broken swoop .\\
{\bf Reinforce}&i don 't know what to do .\\
{\bf MC}&i was just a little bit nervous .\\\hline
{\bf Input}&oatmeal delicious start into a new day\\
{\bf Vanilla-MLE}& i 'll be right back\\
{\bf Vanilla-Sample}&sure if you don 't put it into the water\\
{\bf Reinforce}&i 'm gonna be a little busy with the dishes .\\
{\bf MC}&i 'm gonna make you a little dinner .\\\hline
\end{tabular}
\caption{Appendix: More responses sampled from different models.}
\label{example-appendix2}
\end{table*}

\end{document}